# Building Accurate Semantic Taxonomies from Monolingual MRDs

**German Rigau** and **Horacio Rodríguez**
Departament de LSI.
Universitat Politècnica de Catalunya.
Barcelona. Catalonia.
{g.rigau, horacio}@lsi.upc.es

**Eneko Agirre**
Lengoia eta Informatikoak saila.
Euskal Erriko Universitatea.
Donostia, Basque Country.
jibagbee@si.ehu.es

## Abstract

This paper presents a method that conbines a set of unsupervised algorithms in order to accurately build large taxonomies from any machine-readable dictionary (MRD). Our aim is to profit from conventional MRDs, with no explicit semantic coding. We propose a system that 1) performs fully automatic extraction of taxonomic links from MRD entries and 2) ranks the extracted relations in a way that selective manual refinement is allowed. Tested accuracy can reach around 100% depending on the degree of coverage selected, showing that taxonomy building is not limited to structured dictionaries such as LDOCE.

## 1 Introduction

There is no doubt about the increasing need of owning accurate and broad coverage general lexical/semantic resources for developing NL applications. These resources include Lexicons, Lexical Databases, Lexical Knowledge Bases (LKBs), Ontologies, etc. Many researchers believe that for effective NLP it is necessary to build a LKB which contain class/subclass relations and mechanisms for the inheritance of properties as well as other inferences. The work presented here attempts to lay out some solutions to overcome or alleviate the "lexical bottleneck" problem (Briscoe 91) providing a methodology to build large scale LKBs from conventional dictionaries, in any language. Starting with the seminal work of (Amsler 81) many systems have followed this approach (e.g., Bruce et al. 92; Richardson 97). Why should we propose another one?

Regarding the resources used, we must point out that most of the systems built until now refer to English only and use rather rich, well structured, controlled and explicitly semantically coded dictionaries (e.g. LDOCE 87). This is not the case for most of the available sources for languages other than English. Our aim is to use conventional MRDs, with no explicit semantic coding, to obtain a comparable accuracy.

The system we propose is capable of 1) performing fully automatic extraction (with a counterpart in terms of both recall and precision fall) of taxonomic links of dictionary senses and 2) ranking the extracted relations in a way that selective manual refinement is allowed.

Section 2 shows that applying a conventional pure descriptive approach the resulting taxonomies are not useful for NLP. Our approach is presented in the rest of the paper. Section 3 deals with the automatic selection of the main semantic primitives present in *Diccionario General Ilustrado de la Lengua Española* (DGILE 87), and for each of these, section 4 shows the method for the selection of its most representative genus terms. Section 5 is devoted to the automatic acquisition of large and accurate taxonomies from DGILE. Finally, some conclusions are drawn.

## 2 Acquiring taxonomies from MRDs

A straightforward way to obtain a LKB acquiring taxonomic relations from dictionary definitions can be done following a purely bottom up strategy with the following steps: 1) parsing each definition for obtaining the genus, 2) performing a genus disambiguation procedure, and 3) building a natural classification of the concepts as a concept taxonomy with several tops. Following this purely descriptive methodology, the semantic primitives of the LKB could be obtained by collecting those dictionary senses appearing at the top of the complete taxonomies derived from the dictionary. By characterizing each of these tops, the complete LKB could be produced. For DGILE, the complete noun taxonomy was derived following the automatic method described by (Rigau et al. 97)[1].

---

[1] This taxonomy contains 111,624 dictionary senses and has only 832 dictionary senses which are tops of the taxonomy (these top dictionary senses have no

However, several problems arise a) due to the source (i.e, circularity, errors, inconsistencies, omitted genus, etc.) and b) the limitation of the genus sense disambiguation techniques applied: i.e, (Bruce et al. 92) report 80% accuracy using automatic techniques, while (Rigau et al. 97) report 83%. Furthermore, the top dictionary senses do not usually represent the semantic subsets that the LKB needs to characterize in order to represent useful knowledge for NLP systems. In other words, there is a mismatch between the knowledge directly derived from an MRD and the knowledge needed by a LKB.

To illustrate the problem we are facing, let us suppose we plan to place the FOOD concepts in the LKB. Neither collecting the taxonomies derived from a top dictionary sense (or selecting a subset of the top dictionary senses of DGILE) closest to FOOD concepts (e.g., *substancia* -substance-), nor collecting those subtaxonomies starting from closely related senses (e.g., *bebida* -drinkable liquids- and *alimento* -food-) we are able to collect exactly the FOOD concepts present in the MRD. The first are too general (they would cover non-FOOD concepts) and the second are too specific (they would not cover all FOOD dictionary senses because FOODs are described in many ways).

All these problems can be solved using a mixed methodology. That is, by attaching selected top concepts (and its derived taxonomies) to prescribed semantic primitives represented in the LKB. Thus, first, we prescribe a minimal ontology (represented by the semantic primitives of the LKB) capable of representing the whole lexicon derived from the MRD, and second, following a descriptive approach, we collect, for every semantic primitive placed in the LKB, its subtaxonomies. Finally, those subtaxonomies selected for a semantic primitive are attached to the corresponding LKB semantic category.

Several prescribed sets of semantic primitives have been created as Ontological Knowledge Bases: e.g. Penman Upper Model (Bateman 90), CYC (Lenat & Guha 90), WordNet (Miller 90). Depending on the application and theoretical tendency of the LKB different sets of semantic primitives can be of interest. For instance, WordNet noun top unique beginners are 24 semantic categories. (Yarowsky 92) uses the 1,042 major categories of Roget's thesaurus, (Liddy & Paik 92) use the 124 major subject areas of LDOCE,

(Hearst & Schütze, 95) convert the hierarchical structure of WordNet into a flat system of 726 semantic categories.

In the work presented in this paper we used as semantic primitives the 24 lexicographer's files (or semantic files) into which the 60,557 noun synsets (87,641 nouns) of WordNet 1.5 (WN1.5) are classified[2]. Thus, we considered the 24 semantic tags of WordNet as the main LKB semantic primitives to which all dictionary senses must be attached. In order to overcome the language gap we also used a bilingual Spanish/English dictionary.

## 3 Attaching DGILE dictionary senses to semantic primitives

In order to classify all nominal DGILE senses with respect to WordNet semantic files, we used a similar approach to that suggested by (Yarowsky 92). Rather than collect evidence from a blurred corpus (words belonging to a Roget's category are used as seeds to collect a subcorpus for that category; that is, a window context produced by a seed can be placed in several subcorpora), we collected evidence from dictionary senses labelled by a conceptual distance method (that is, a definition is placed in one semantic file only).

This task is divided into three fully automatic consecutive subtasks. First, we tag a subset (due to the difference in size between the monolingual and the bilingual dictionaries) of DGILE dictionary senses by means of a process that uses the conceptual distance formula; second, we collect salient words for each semantic file; and third, we enrich each DGILE dictionary sense with a semantic tag collecting evidence from the salient words previously computed.

### 3.1 Attach WordNet synsets to DGILE headwords.

For each DGILE definition, the conceptual distance between headword and genus has been computed using WN1.5 as a semantic net. We obtained results only for those definitions having English translations for both headword and genus. By computing the conceptual distance between two words ($w_1$,$w_2$) we are also selecting those concepts ($c_{1i}$,$c_{2j}$) which represent them and seem to be closer with respect to the semantic net

---

hypernyms), and 89,458 leaves (which have no hyponyms). That is, 21,334 definitions are placed between the top nodes and the leaves.

[2]One could use other semantic classifications because using this methodology a minimal set of informed seeds are needed. These seeds can be collected from MRDs, thesauri or even by introspection, see (Yarowsky 95).

used. Conceptual distance is computed using formula (1).

$$(1) \quad dist(w_1, w_2) = \min_{\substack{c_{1_i} \in w_1 \\ c_{2_i} \in w_2}} \sum_{c_k \in path(c_{1_i}, c_{2_i})} \frac{1}{depth(c_k)}$$

That is, the conceptual distance between two concepts depends on the length of the shortest path[3] that connects them and the specificity of the concepts in the path.

| | |
|---|---|
| Noun definitions | 93,394 |
| Noun definitions with genus | 92,693 |
| Genus terms | 14,131 |
| Genus terms with bilingual translation | 7,610 |
| Genus terms with WN1.5 translation | 7,319 |
| Headwords | 53,455 |
| Headwords with bilingual translation | 11,407 |
| Headwords with WN1.5 translation | 10,667 |
| Definitions with bilingual translation | 30,446 |
| Definitions with WN1.5 translation | 28,995 |

Table 1, data of first attachment using conceptual distance.

As the bilingual dictionary is not disambiguated with respect to WordNet synsets (every Spanish word has been assigned to all possible connections to WordNet synsets), the degree of polysemy has increased from 1.22 (WN1.5) to 5.02, and obviously, many of these connections are not correct. This is one of the reasons why after processing the whole dictionary we obtained only an accuracy of 61% at a sense (synset) level (that is, correct synsets attached to Spanish headwords and genus terms) and 64% at a file level (that is, correct WN1.5 lexicographer's file assigned to DGILE dictionary senses)[4]. We processed 32,208[5] dictionary definitions, obtaining 29,205 with a synset assigned to the genus (for the rest we did not obtain a bilingual-WordNet relation between the headword and the genus, see Table 1).

In this way, we obtained a preliminary version of 29,205 dictionary definitions semantically labelled (that is, with Wordnet lexicographer's files) with an accuracy of 64%. That is, a corpus (collection of dictionary senses) classified in 24 partitions (each one corresponding to a semantic category). Table 2 compares the distribution of these DGILE dictionary senses (see column a) with respect to WordNet semantic categories. The greatest differences appear with the classes ANIMAL and PLANT, which correspond to large taxonomic scientific classifications occurring in WN1.5 but which do not usually appear in a bilingual dictionary.

---

[3] We only consider hypo/hypermym relations.
[4] To evaluate this process, we select at random a test set with 391 noun senses that give a confidence rate of 95%.
[5] The difference with 30,446 is accounted for by repeated headword and genus for an entry.

### 3.2 Collect the salient words for every semantic primitive.

Once we have obtained the first DGILE version with semantically labelled definitions, we can collect the salient words (that is, those representative words for a particular category) using a Mutual Information-like formula (2), where w means word and SC semantic class.

$$(2) \quad AR(w, SC) = \Pr(w|SC) \log_2 \frac{\Pr(w|SC)}{\Pr(w)}$$

Intuitively, a salient word[6] appears significantly more often in the context of a semantic category than at other points in the whole corpus, and hence is a better than average indicator for that semantic category. The words selected are those most relevant to the semantic category, where relevance is defined as the product of salience and local frequency. That is to say, important words should be distinctive and frequent.

We performed the training process considering only the content word forms from dictionary definitions and we discarded those salient words with a negative score. Thus, we derived a lexicon of 23,418 salient words (one word can be a salient word for many semantic categories, see Table 2, columns b and c).

### 3.3 Enrich DGILE definitions with WordNet semantic primitives.

Using the salient words per category (or semantic class) gathered in the previous step we labelled the DGILE dictionary definitions again.

When any of the salient words appears in a definition, there is evidence that the word belongs to the category indicated. If several of these words appear, the evidence grows.

---

[6] Instead of word lemmas, this study has been carried out using word forms because word forms rather than lemmas are representative of typical usages of the sublanguage used in dictionaries.

| Semantic file | #DGILE senses (a) | #Content words(b) | #Salient words(c) | #DGILE senses (d) | #WordNet synsets |
|---|---|---|---|---|---|
| 03 tops | 77 (0.2%) | 540 | - | - | 35 (0.0%) |
| 04 act | 3,138 (10.7%) | 16,963 | 2,593 | 4,188 (4.8%) | 4895 (8.0%) |
| 05 animal | 712 (2.4%) | 6,191 | 849 | 4,544 (5.2%) | 7,112 (11.7%) |
| 06 artifact | 6,915 (23.7%) | 45,988 | 4,515 | 12,958 (14.9%) | 9,101 (15.0%) |
| 07 attribute | 2,078 (7.1%) | 11,069 | 1,571 | 4,146 (4.8%) | 2,526 (4.2%) |
| 08 body | 621 (2.1%) | 4,285 | 665 | 3,208 (3.6%) | 1,370 (2.3%) |
| 09 cognition | 1,556 (5.3%) | 9,699 | 1,362 | 3,672 (4.2%) | 2,007 (3.3%) |
| 10 communication | 4,076 (13.9%) | 24,633 | 3,301 | 6,012 (6.9%) | 4,115 (6.8%) |
| 11 event | 541 (1.8%) | 3,071 | 477 | 1,544 (1.7%) | 752 (1.2%) |
| 12 feeling | 306 (1.0%) | 1,623 | 263 | 1,016 (1.2%) | 397 (0.6%) |
| 13 food | 749 (2.5%) | 4,679 | 717 | 2,614 (3.0%) | 2,290 (3.8%) |
| 14 group | 661 (2.2%) | 4,338 | 647 | 3,074 (3.5%) | 1,661 (2.7%) |
| 15 place | 416 (1.4%) | 2,587 | 402 | 2,073 (2.4%) | 1,755 (2.9%) |
| 16 motive | 15 (0.0%) | 87 | 9 | 22 (0.0%) | 28 (0.0%) |
| 17 object | 437 (1.5%) | 2,733 | 412 | 1,645 (1.9%) | 839 (1.4%) |
| 18 person | 3,279 (11.2%) | 19,273 | 2,304 | 13,901 (16.0%) | 5,563 (9.1%) |
| 19 phenomenon | 147 (0.5%) | 784 | 114 | 425 (0.4%) | 452 (0.7%) |
| 20 plant | 581 (2.0%) | 4,965 | 700 | 4,234 (4.9%) | 7,971 (13.2%) |
| 21 possession | 287 (1.0%) | 1,712 | 278 | 1,033 (1.2%) | 829 (1.4%) |
| 22 process | 211 (0.7%) | 987 | 177 | 6948 (8.0%) | 445 (0.7%) |
| 23 quantity | 344 (1.2%) | 2,179 | 317 | 1,502 (1.7%) | 1,050 (1.7%) |
| 24 relation | 102 (0.3%) | 600 | 76 | 288 (0.3%) | 343 (0.6%) |
| 25 shape | 165 (0.6%) | 1,040 | 172 | 677 (0.8%) | 284 (0.4%) |
| 26 state | 805 (2.7%) | 4,469 | 712 | 1,973 (2.3%) | 1,870 (3.0%) |
| 27 substance | 642 (2.2%) | 5,002 | 734 | 3,518 (4.0%) | 2,068 (3.4%) |
| 28 time | 344 (1.2%) | 2,172 | 321 | 1,544 (1.8%) | 799 (1.3%) |
| Total | 32,208 | 181,669 | 23,418 | 82,759 | 60,557 |

Table 2, comparison of the two labelling process (and salient words per context) with to respect WN1.5 semantic tags.

We add together their weights, over all words in the definition, and determine the category for which the sum is greatest, using formula (3).

$$(3) \quad W(SC) = \sum_{w \in definition} AR(w, SC)$$

Thus, we obtained a second semantically labelled version of DGILE (see table 2, column d). This version has 86,759 labelled definitions (covering more than 93% of all noun definitions) with an accuracy rate of 80% (we have gained, since the previous labelled version, 62% coverage and 16% accuracy).

The main differences appear (apart from the classes ANIMAL and PLANT) in the classes ACT and PROCESS. This is because during the first automatic labelling many dictionary definitions with genus *acción* (act or action) or *efecto* (effect) were classified erroneously as ACT or PROCESS.

These results are difficult to compare with those of [Yarowsky 92]. We are using a smaller context window (the noun dictionary definitions have 9.68 words on average) and a microcorpus (181,669 words). By training salient words from a labelled dictionary (only 64% correct) rather than a raw corpus we expected to obtain less noise.

Although we used the 24 lexicographer's files of WordNet as semantic primitives, a more fine-grained classification could be made. For example, all FOOD synsets are classified under **<food, nutrient>** synset in file 13. However, FOOD concepts are themselves classified into 11 subclasses (i.e., **<yolk>**, **<gastronomy>**, **<comestible, edible, eatable, ...>**, etc.). Thus, if the LKB we are planning to build needs to represent **<beverage, drink, potable>** separately from the concepts **<comestible, edible, eatable, ...>** a finer set of semantic primitives should be chosen, for instance, considering each direct hyponym of a synset belonging to a semantic file also as a new semantic primitive or even selecting for each semantic file the level of abstraction we need.

A further experiment could be to iterate the process by collecting from the second labelled dictionary (a bigger corpus) a new set of salient words and reestimating again the semantic tags for all dictionary senses (a similar approach is used in Riloff & Shepherd 97).

## 4 Selecting the main top beginners for a semantic primitive

This section is devoted to the location of the main top dictionary sense taxonomies for a given semantic primitive in order to correctly attach all these taxonomies to the correct semantic primitive in the LKB.

In order to illustrate this process we will locate the main top beginners for the FOOD dictionary senses. However, we must consider that many of these top beginners are structured. That is, some of them belong to taxonomies derived from other ones, and then cannot be directly placed within the FOOD type. This is the case of *vino* (*wine*), which is a *zumo* (*juice*). Both are top beginners for FOOD and one is a hyponym of the other.

First, we collect all genus terms from the whole set of DGILE dictionary senses labelled in the previous section with the FOOD tag (2,614 senses), producing a lexicon of 958 different genus terms (only 309, 32%, appear more than once in the FOOD subset of dictionary senses[7]).

As the automatic dictionary sense labelling is not free of errors (around 80% accuracy)[8] we can discard some senses by using filtering criteria.

• **Filter 1** (F1) removes all FOOD genus terms not assigned to the FOOD semantic file during the mapping process between the bilingual dictionary and WordNet.

• **Filter 2** (F2) selects only those genus terms which appear more times as genus terms in the FOOD category. That is, those genus terms which appear more frequently in dictionary definitions belonging to other semantic tags are discarded.

• **Filter 3** (F3) discards those genus terms which appear with a low frequency as genus terms in the FOOD semantic category. That is, infrequent genus terms (given a certain threshold) are removed. Thus, F3>1 means that the filtering criteria have discarded those genus terms

---

[7] We select this group of genus for the test set.
[8] Most of them are not really errors. For instance, all fishes must be ANIMALs, but some of them are edible (that is, FOODs). Nevertheless, all fishes labelled as FOOD have been considered mistakes.

appearing in the FOOD subset of dictionary definitions less than twice.

Table 4 shows the first 10 top beginners for FOOD. Bold face is used for those genus terms removed by filter 2. Thus, *pez* -fish- is an ANIMAL.

| 90 | bebida (drink) | 48 | pasta (pasta, etc.) |
|----|----------------|----|---------------------|
| 86 | vino (wine)    | 40 | pan (bread)         |
| 78 | **pez (fish)** | 39 | plato (dish)        |
| 56 | comida (food)  | 33 | guisado (casserole) |
| 55 | carne (meat)   | 32 | salsa (souce)       |

Table 4, frequency of main top beginners for FOOD.

Table 5 shows the performance of the second labelling with respect to filter 3 (genus frequency) varying the threshold. From left to right, filter, number of genus terms selected (#GT), accuracy (A), number of definitions (#D) and their respective accuracy.

| LABEL2 + F3 | #GT | A | #D | A |
|-------------|-----|-----|-------|-----|
| F3>9 | 32 | 89% | 908 | 88% |
| F3>8 | 37 | **90%** | 953 | 88% |
| F3>7 | 39 | 88% | 969 | 87% |
| F3>6 | 45 | 88% | 1,011 | 87% |
| F3>5 | 51 | 87% | 1,047 | 82% |
| F3>4 | 62 | 85% | 1,102 | 86% |
| F3>3 | 73 | 78% | 1,146 | 84% |
| F3>2 | 99 | 69% | 1,224 | 80% |
| F3>1 | 151 | 62% | 1,328 | 77% |

Table 5. performance of filter 3.

| LABEL2 + F1 | #GT | A | #D | A |
|-------------|-----|-----|-------|-----|
| F1+F3>9 | 31 | **94%** | 895 | **90%** |
| F1+F3>8 | 35 | **95%** | 931 | **90%** |
| F1+F3>7 | 37 | **91%** | 947 | 89% |
| F1+F3>6 | 43 | **92%** | 989 | **90%** |
| F1+F3>5 | 49 | **92%** | 1,025 | **90%** |
| F1+F3>4 | 55 | **91%** | 1,055 | **90%** |
| F1+F3>3 | 64 | 85% | 1,091 | 88% |
| F1+F3>2 | 85 | 82% | 1,152 | 87% |
| F1+F3>1 | 125 | 78% | 1,234 | 86% |

Table 6, performance of filter 1 varying filter 3.

Tables 6 and 7 show that at the same level of genus frequency, filter 2 (removing genus terms which are more frequent in other semantic categories) is more accurate that filter 1 (removing all genus terms the translation of which cannot be FOOD). For instance, no error appears when selecting those genus terms which appear 10 or more times (F3) and are more frequent in that category than in any other (F2).

Table 8 shows the coverage of correct genus terms selected by criteria F1 and F2 to respect criteria F3. Thus, for genus terms appearing 10 or more times, by using either of the two criteria we are collecting 97% of the correct ones. That is, in both cases the criteria discards less than 3% of correct genus terms.

| LABEL2 + F2 | #GT | A | #D | A |
|---|---|---|---|---|
| F2+F3>9 | 31 | **100%** | 893 | **100%** |
| F2+F3>8 | 35 | **100%** | 929 | **100%** |
| F2+F3>7 | 37 | **95%** | 945 | **98%** |
| F2+F3>6 | 41 | **94%** | 973 | **98%** |
| F2+F3>5 | 47 | **92%** | 1,009 | **97%** |
| F2+F3>4 | 56 | **91%** | 1,054 | **96%** |
| F2+F3>3 | 65 | **87%** | 1,090 | **95%** |
| F2+F3>2 | 82 | **83%** | 1,141 | **93%** |
| F2+F3>1 | 123 | **82%** | 1,223 | **92%** |

Table 7, performance of filter 2 varying filter 3.

|  | Coverage vs F1 | Coverage vs F2 |
|---|---|---|
| F3>9 | **97%** | **97%** |
| F3>8 | **95%** | **95%** |
| F3>7 | **95%** | **95%** |
| F3>6 | **96%** | **91%** |
| F3>5 | **96%** | **92%** |
| F3>4 | **89%** | **90%** |
| F3>3 | **90%** | **89%** |
| F3>2 | **86%** | **83%** |
| F3>1 | **83%** | **81%** |

Table 8, coverage of second labelling with respect to filter 1 and 2 varying filter 3.

## 5 Building automatically large scale taxonomies from DGILE

The automatic Genus Sense Disambiguation task in DGILE has been performed following (Rigau et al. 97). This method reports 83% accuracy when selecting the correct hypernym by combining eight different heuristics using several methods and types of knowledge. Using this combined technique the selection of the correct hypernym from DGILE had better performance than those reported by (Bruce et al. 92) using LDOCE.

Once the main top beginners (relevant genus terms) of a semantic category are selected and every dictionary definition has been disambiguated, we collect all those pairs labelled with the semantic category we are working on having one of the genus terms selected. Using these pairs we finally build up the complete taxonomy for a given semantic primitive. That is, in order to build the complete taxonomy for a semantic primitive we fit the lower senses using the second labelled lexicon and the genus selected from this labelled lexicon.

Table 9 summarizes the sizes of the FOOD taxonomies acquired from DGILE with respect to filtering criteria and the results manually obtained by (Castellón 93)[9] where 1) is (Castellón 93), (2) F2 + F3 > 9 and (3) F2 + F3 > 4.

| FOOD | (1) | (2) | (3) |
|---|---|---|---|
| Genus terms | 62 | 33 | 68 |
| Dictionary senses | 392 | 952 | 1,242 |
| Levels | 6 | 5 | 6 |
| Senses in level 1 | 2 | 18 | 48 |
| Senses in level 2 | 67 | 490 | 604 |
| Senses in level 3 | 88 | 379 | 452 |
| Senses in level 4 | 67 | 44 | 65 |
| Senses in level 5 | 87 | 21 | 60 |
| Senses in level 6 | 6 | 0 | 13 |

Table 9, comparison of FOOD taxonomies.

Using the first set of criteria (F2+F3>9), we acquire a FOOD taxonomy with 952 senses (more than two times larger than if it is done manually). Using the second one (F2+F3>4), we obtain another taxonomy with 1,242 (more than three times larger). While using the first set of criteria, the 33 genus terms selected produce a taxonomic structure with only 18 top beginners, the second set, with 68 possible genus terms, produces another taxonomy with 48 top beginners. However, both final taxonomic structures produce more flat taxonomies than if the task is done manually. This is because we are restricting the inner taxonomic genus terms to those selected by the criteria (33 and 68 respectively). Consider the following taxonomic chain, obtained in a semiautomatic way by (Castellón 93):

bebida_1_3 <- **líquido**_1_6 <- zumo_1_1 <- vino_1_1 <- rueda_1_1

As *líquido* -liquid- was not selected as a possible genus (by the criteria described above), the taxonomic chain for that sense is:

zumo_1_1 <- vino_1_1 <- rueda_1_1

---

[9] We used the results reported by (Castellón 93) as a baseline because her work was done using the same Spanish dictionary.

Thus, a few arrangements (18 or 48 depending on the criteria selected) must be done at the top level of the automatic taxonomies. Studying the main top beginners we can easily discover an internal structure between them. For instance, placing all *zumo* (*juice*) senses within *bebida* (*drink*).

Performing the same process for the whole dictionary we obtained for F2+F3>9 a taxonomic structure of 35,099 definitions and for F2+F3>4 the size grows to 40,754.

## 6 Conclusions

We proposed a novel methodology which combines several structured lexical knowledge resources for acquiring the most important genus terms of a monolingual dictionary for a given semantic primitive. Our approach for building LKBs is mainly descriptive (the main source of knowledge is MRDs), but a minimal prescribed structure is provided (the semantic primitives of the LKB). Using the most relevant genus terms for a particular semantic primitive and applying a filtering process, we presented a method to construct fully automatically taxonomies from any conventional dictionary. This approach differs from previous ones because we are considering senses as lexical units of the LKB (e.g., in contrast to Richardson 97 who links words) and the mixed methodology applied (e.g, the complete descriptive approach of Bruce et al. 92).

The results show that the construction of taxonomies using lexical resources is not limited to highly structured MRDs. Applying appropriate techniques, conventional dictionaries such as DGILE could be useful resources for building automatically substantial pieces of an LKB .

## Acknowledgments

This research has been partially funded by the Spanish Research Department (ITEM Project TIC96-1243-C03-03), the Catalan Research Department (CREL project), and the UE Comision (EuroWordNet LE4003).